\documentclass[10pt, a4paper]{article}
\usepackage{lrec2022} 
\usepackage{multibib}
\newcites{languageresource}{Language Resources}
\usepackage{graphicx}
\usepackage{tabularx}
\usepackage{soul}

\usepackage{textcomp}
\usepackage{titlesec}
\titleformat{\section}{\normalfont\large\bfseries\center}{\thesection.}{1em}{}
\titleformat{\subsection}{\normalfont\SmallTitleFont\bfseries\raggedright}{\thesubsection.}{1em}{}
\titleformat{\subsubsection}{\normalfont\normalsize\bfseries\raggedright}{\thesubsubsection.}{1em}{}
\renewcommand\thesection{\arabic{section}}
\renewcommand\thesubsection{\thesection.\arabic{subsection}}
\renewcommand\thesubsubsection{\thesubsection.\arabic{subsubsection}}

\usepackage{epstopdf}
\usepackage[utf8]{inputenc}

\usepackage{hyperref}
\usepackage{xstring}

\usepackage{color}

\title{PerPaDa: A Persian Paraphrase Dataset based on Implicit Crowdsourcing Data Collection}

\name{Salar Mohtaj\textsuperscript{1,2}, Fatemeh Tavakkoli\textsuperscript{3}, Habibollah Asghari\textsuperscript{4}} 

\address{\textsuperscript{1} Technische Universit\"at Berlin, Berlin, Germany \\
        \textsuperscript{2} German Research Centre for Artificial Intelligence (DFKI), Projektb\"uro Berlin, Germany \\
        \textsuperscript{3} Freie Universit\"at Berlin, Berlin, Germany \\
        \textsuperscript{4} ICT Research Institute of ACECR, Tehran, Iran \\
         salar.mohtaj@tu-berlin.de, f.tavakkoli@fu-berlin.de, habib.asghari@ictrc.ac.ir}

\abstract{
In this paper we introduce PerPaDa, a Persian paraphrase dataset that is collected from users' input in a plagiarism detection system. As an implicit crowdsourcing experience, we have gathered a large collection of original and paraphrased sentences from Hamtajoo; a Persian plagiarism detection system, in which users try to conceal cases of text re-use in their documents by paraphrasing and re-submitting manuscripts for analysis. The compiled dataset contains 2446 instances of paraphrasing. In order to improve the overall quality of the collected data, some heuristics have been used to exclude sentences that don't meet the proposed criteria. The introduced corpus is much larger than the available datasets for the task of paraphrase identification in Persian. Moreover, there is less bias in the data compared to the similar datasets, since the users did not try some fixed predefined rules in order to generate similar texts to their original inputs. 
 \\ \newline \Keywords{Paraphrase, Persian, Implicit crowdsourcing, Plagiarism detection } }

\begin{document}

\maketitleabstract

\section{Introduction}
\label{sec:intro}

Paraphrase identification is the task of investigating if a piece of text is a paraphrase of another one~\cite{DBLP:conf/icbk/HuntDZGOJKKSZOW19}. It is a basic task for different Natural Language Processing (NLP) tasks like Plagiarism Detection (PD) and question answering. Paraphrase identification could improve the overall performance of these tasks by enabling systems to detect semantically similar or related sentences or phrases. For instance, a plagiarism detection system that empowered by a  paraphrase identification module not only can detect cases of verbatim text re-use, but also the cases in which people try to conceal plagiarism by using different wordings and the other paraphrasing techniques. \par
There are different approaches for identification of paraphrases. A possible approach for measuring if two sentences are semantically similar (i.e., paraphrased) is two measure cosine similarity between two sentences~\cite{DBLP:conf/paclic/MahmoudZ17}. However, more recent approaches are using Machine Learning (ML) models to train a classifier to detect if two pieces of text are paraphrased. Moreover, pre-trained language models (e.g., BERT~\cite{DBLP:conf/naacl/DevlinCLT19}) have been using recently for the task of paraphrase identification~\cite{DBLP:conf/iclr/0225BYWXBPS20}. The neural networks could also be used to generate paraphrases from textual data~\cite{DBLP:conf/jcdl/WahleRMG21}. \par
A training dataset is an essential part of all supervised learning approaches in ML to train a model. A paraphrase identification dataset is an important piece of puzzle to train an outstanding paraphrase detection model. In this paper we introduced \textit{PerPaDa}, a new \underline{Per}sian \underline{Pa}raphrase \underline{Da}taset in Persian. The data is collected from users' input to Hamtajoo\footnote{www.hamtajoo.ir} plagiarism detection system. Hamtajoo is a Persian plagiarism detection tool that is being used by journals, editorial board, conferences, faculty members and students to detect cases of inadvertent or intentional text re-use in scientific papers. \par
As an implicit crowdsourcing experience, we have gathered a large collection of original and paraphrased sentences from Hamtajoo, when users try to conceal cases of text re-use in their documents by paraphrasing and re-submitting manuscripts for analysis. 
The proposed dataset contains 2446 instances of paraphrased sentences. Based on our knowledge on the current Persian paraphrase detection dataset, \textit{PerPaDa} is much larger than the available datasets. 
Bias is an important issue in experiments which try to use crowdsourcing for data curation,
annotation and evaluation for ML and NLP \nocite{DBLP:conf/wsdm/Eickhoff18}. Similarly, it is a drawback of the available paraphrase datasets based on crowdsourcing that crowd-workers intend to follow some instructions which usually are provided by the experiment designers. This leads to a bias on the applied strategies by people to paraphrase original text. We believe that the implicit crowdsourcing approach that has been used in this paper results to a more diverse and general dataset which covers different paraphrasing strategies. \par 
The paper is organized as follow; Section \ref{sec:relatedwork} contains a brief overview of some of the Persian and English corpora for the task of paraphrase identification. We explain the raw data collection procedure from Hamtajoo platform in Section \ref{sec:datacollection}. The main steps to construct the \textit{PerPaDa} dataset are described in Section \ref{sec:corpusconstruction}. We represent some statistics on the data and the evaluation of \textit{PerPaDa} in Section \ref{sec:evaluation} and finally conclude the paper in Section \ref{sec:conclusion}.

\section{Related Work}
\label{sec:relatedwork}
In this section, some of the related datasets for the task of paraphrase identification will be introduced. \par
Although the paraphrasing corpora are very limited in Persian, a number of datasets for the task of text re-use detection (i.e., plagiarism detection) have been introduced in recent years. Usually they include paraphrased or obfuscated text that are inserted from source documents into suspicious ones, and PD systems should automatically detect the position of these re-used pieces of text in the source and suspicious documents. We will review some of them in this section. \par
Khoshnavataher et. al. compiled a Persian plagiarism detection based on automatically generated cases of paraphrasing~\cite{DBLP:conf/clef/KhoshnavataherZ15}. these automated approaches include shuffling words in sentences, substitution of some words with their synonyms and addition/deletion of some words to/from a sentence. \par
Asghari et. al. from the same team tried to enrich the previous data by incorporating manually paraphrased sentences~\cite{DBLP:journals/jodl/AsghariFMF21}. In addition to the mentioned automated approaches to generate obfuscated sentences, they also used more than 150 pairs of paraphrased sentences based on a crowdsourcing experiment. \par

Mashhadirajab et. al. proposed a text alignment corpus for Persian plagiarism detection, which includes more than 11000 documents and about 11600 plagiarism cases in PAN format~\cite{DBLP:conf/fire/MashhadirajabSA16}. They simulated different types of plagiarism by exploiting manually, semi-automatically, or automatically approaches to generate paraphrases in this large-scale corpus~\cite{DBLP:conf/fire/MashhadirajabSA16}. \par
To best of our knowledge, there is no Persian paraphrased identification dataset that is compiled for this standalone task. \textit{PerPaDa} could be the first such a dataset that can be used to train Persian paraphrase generation and identification models. 

\section{Data Collection}
\label{sec:datacollection}
In this section we introduce the data collection procedure, in which we've collected the source data to compile \textit{PerPaDa} dataset. 

\subsection{Hamtajoo plagiarism detection system}
Hamtajoo is a Persian plagiarism detection system for investigating patterns of text re-use in Persian academic papers~\cite{DBLP:journals/corr/abs-2112-13742}. The system works on document level at the first stage and then focuses on paragraph and sentence level in the second detailed comparison stage. The system was officially introduced on 2017 and has been using by academicians to prevent and detect cases of text matching since then. \par
\begin{figure}[!h]
\begin{center}
\includegraphics[width=0.46\textwidth]{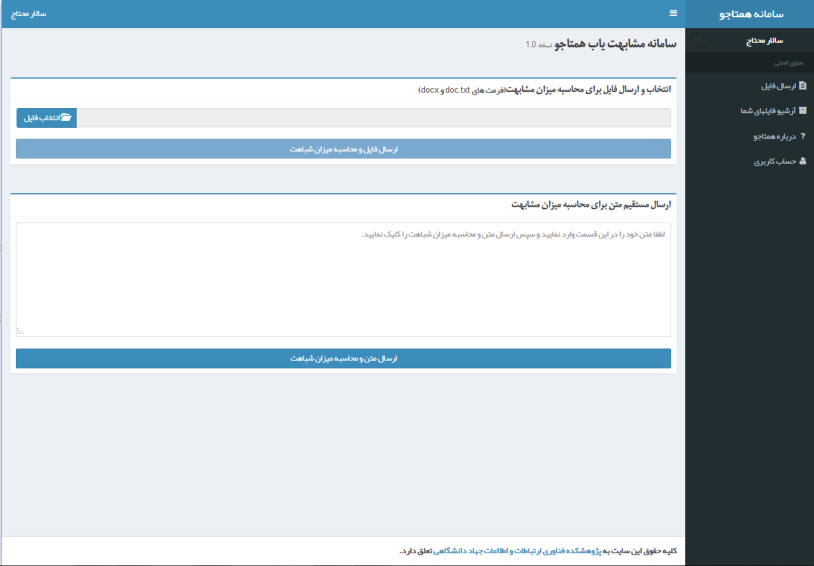}
\caption{The submission page of Hamtajoo system (the menu and text are in Persian) \protect\cite{DBLP:journals/corr/abs-2112-13742}}
\label{fig:1}
\end{center}
\end{figure}
\begin{figure}
\begin{center}
\includegraphics[width=0.46\textwidth]{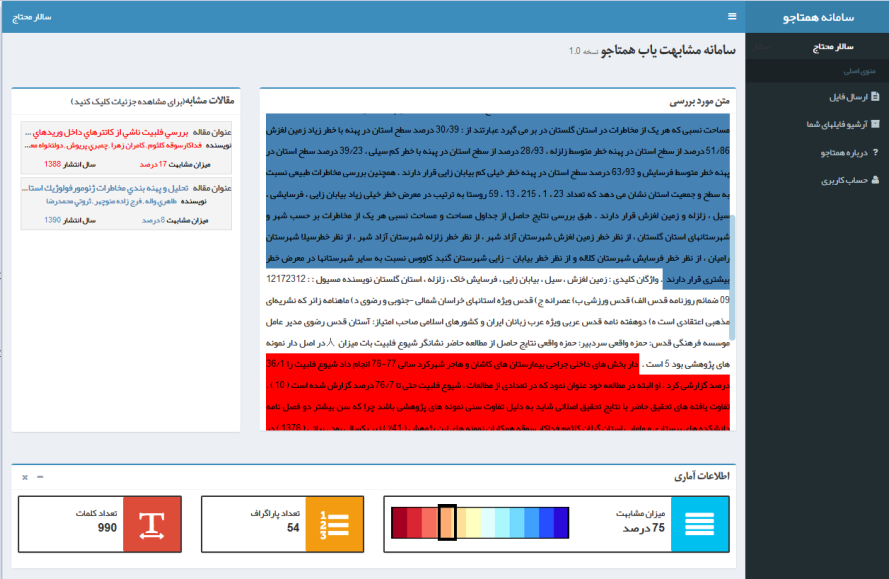}
\caption{The Result page of Hamtajoo system (the menu and text are in Persian) \protect\cite{DBLP:journals/corr/abs-2112-13742}.}
\label{fig:2}
\end{center}
\end{figure}

Two screenshots of Hamtajoo platform are shown in Figures \ref{fig:1} and \ref{fig:2}. The first figure shows the submission page of the platform, where users can either upload their textual document or directly insert text into related the field.

Figure \ref{fig:2} shows the Result page, where the sections with a piece of re-used text are highlighted. The origin of the re-used texts are also represented in this page, so users can track reasons behind the system's decisions.

There are two main use cases of Hamtajoo and the other plagiarism detection tools for the end users; The tool could be used by students, journal editors and university faculties as dictated by the workflow before publishing papers or theses. On the other hand, these systems are sometimes used to detect potential text re-use cases to reduce or conceal those text matching cases by paraphrasing or removing suspicious parts from manuscripts. \par
As an implicit crowdsourcing study, in this research we targeted the second use case of Hamtajoo, where users try to use the system to detect plagiarism cases and then conceal it by paraphrasing. We believe that the resulting pairs of original and paraphrased pieces of text are a rich resource to train paraphrase identification models because:
\begin{enumerate}
    \item There is no bias in the data. In other words, users don't follow any instruction to generate paraphrases. 
    \item The dataset is huge, since the system is widely used by academicians in Iran.
    \item It covers different scientific topics and domains including humanity, engineering etc.
\end{enumerate}

\subsection{Raw data gathering}
\begin{figure}[!h]
\begin{center}
\includegraphics[width=0.46\textwidth]{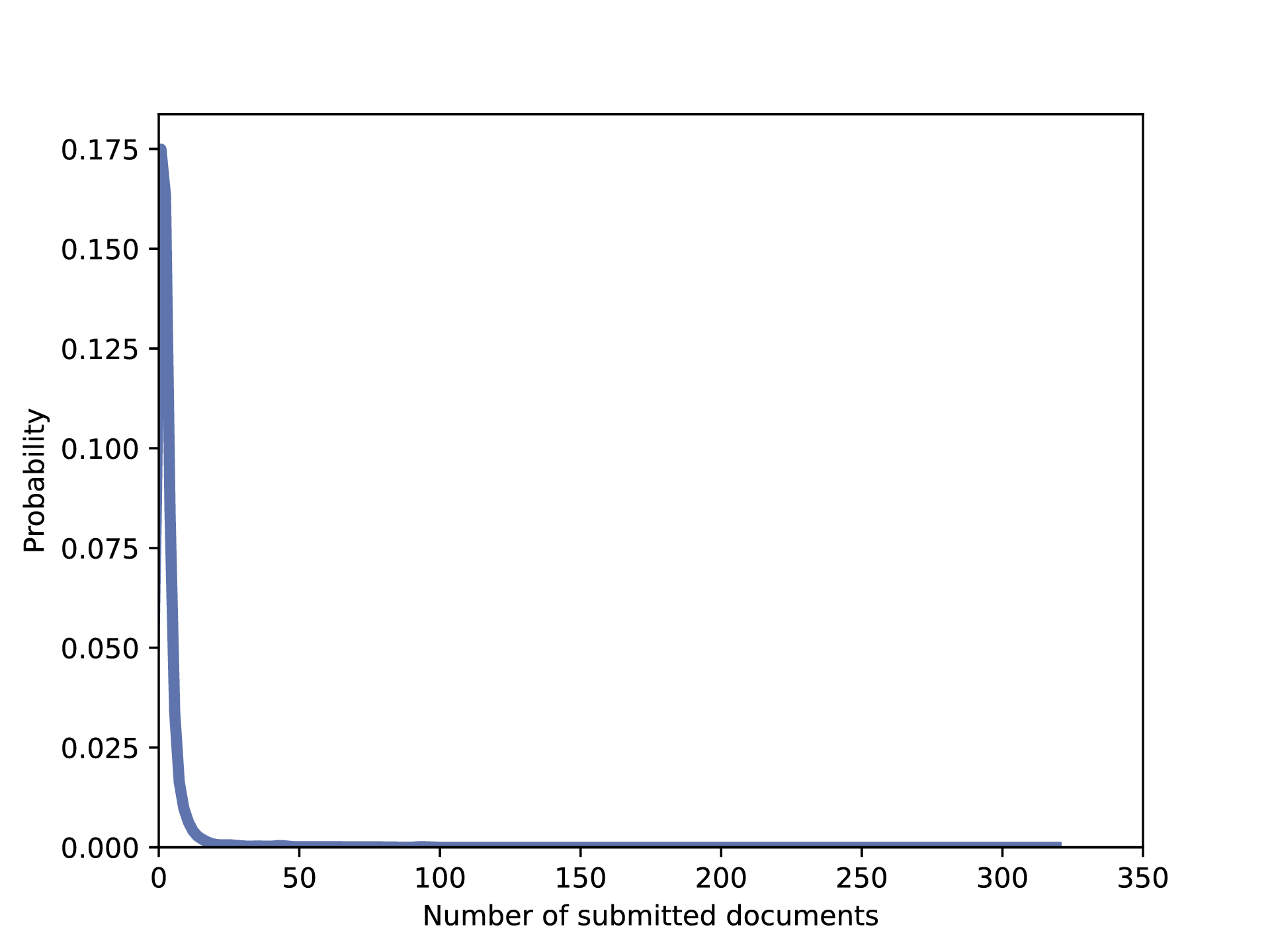}
\caption{The distribution of the number of submitted documents for plagiarism detection by Hamtajoo users}
\label{fig:3}
\end{center}
\end{figure}

As mentioned earlier, we have focused on those use cases in which users employ the plagiarism detection system to find case of text re-use and then try to conceal them by paraphrasing. For this purpose, we excluded organizational accounts of the system. These users are mainly journal editors who use the system to check manuscripts for plagiarism before publication. For the remaining individual users, Figure \ref{fig:3} shows the distribution of number of checked documents by the system.

The number of submitted documents by individual users shows a long tail distribution. Most of the users (i.e., almost 75\%) uploaded fewer than three documents for checking against plagiarism. On the other hand, there are a few users who submitted more than 300 documents into the system. \par
\begin{figure}[!h]
\begin{center}
\includegraphics[width=0.46\textwidth]{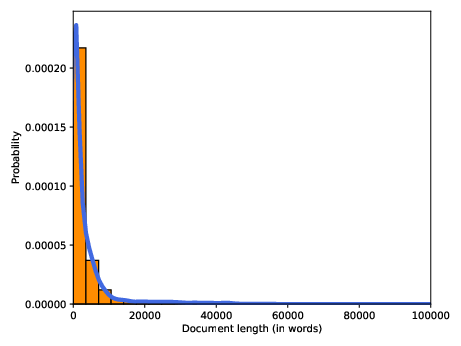}
\caption{The histogram and length distribution of the target documents}
\label{fig:4}
\end{center}
\end{figure}

Since the idea is to compare multiple submission of a document to extract those parts which paraphrased by users, we excluded the users who submitted just one document in the system. The length distribution of the reminded \textbf{18111} documents is shown in Figure \ref{fig:4}. The length distribution follow a power law distribution too. Here again, there are too many documents which are shorter than 5,000 words and a few documents that are longer than 80,000 words. The detailed steps to extract paraphrased sentences from these documents are explained in the succeeding section.

\section{Corpus Construction}
\label{sec:corpusconstruction}
In this section we elaborate the main steps to extract paraphrased sentences out of initial selected documents. \par
\begin{figure*}
\begin{center}
\includegraphics[width=0.5\textwidth]{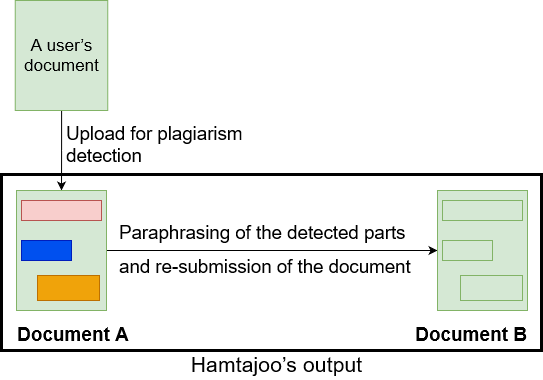}
\caption{A sample scenario in which a user upload a document, and then re-submit it after paraphrasing to conceal cases of text matching}
\label{fig:5}
\end{center}
\end{figure*}

A sample scenario in which a user use the plagiarism detection system to find case of text re-use and then try to conceal them by paraphrasing is depicted in Figure \ref{fig:5}. As it's highlighted in the image, a user would try to re-write those colored parts in document A and re-submit the modified document to the system to check if it can detect those paraphrased parts (Document B). Regardless of the final results (i.e., whether the system can detect those paraphrased sections), the matching between the original sentences in \textit{Document A} and the re-written ones in \textit{Document B} is a valuable resource of paraphrased sentences. It should be noted that a user may re-submit different versions of the initial document after paraphrasing various parts of the paper, and this matching could be repeated for each pair. \par
The main steps for matching the original detected sentences and the paraphrased ones are as follow:
\begin{enumerate}
    \item Detection of near duplicate documents for each user.
    \item Ordering documents in the near duplicate clusters based on the time of submission. 
    \item Extracting detected sentences in the \textit{lead} documents.
    \item Searching for the \textit{paraphrased} sentences at the similar position in the \textit{subsequent} documents.
    \item Post-processing of the extracted pairs by applying some heuristics to exclude low quality pairs.
\end{enumerate}
These steps are elaborated in the succeeding sub-sections.

\subsection{Near Duplicate Detection}
In this step all the submitted documents by a user are clustered into the group of near duplicate documents. Since users may upload different documents with totally different contents, this step help us to screen those documents that potentially includes cases of paraphrasing for the use in the next steps. \par
For measuring the similarity between submitted documents by a user, we computed cosine similarity between \textit{TF-IDF} vectors of pairs of documents. We set the similarity threshold to [0.9 - 1). So those documents that have cosine similarity greater or equal to 0.9 and lower than 1 are considered as near duplicate documents. The lower band threshold is a heuristic that comes after doing a number of experiments. We don't consider documents with the cosine similarly of 1 (i.e., exact match) for further analysis because they definitely don't contain cases of paraphrasing. The reason of exact match documents in the system could be the lack of getting response from plagiarism detection of a document in time which leads to re-submission of the same document once more into the system. \par
Among the clusters of documents of each user, we kept those groups that include at least two documents for further analysis in the next steps. 

\subsection{Ordering the Documents}
In this tiny step, the near duplicate documents are ordered based their submission date/time. The reason is to be able to track changes on those documents that submitted later (i.e., re-submissions). So, the comparison of the documents is always have been done forward and the \textit{lead} documents compared to those submitted later (i.e., \textit{subsequent} documents).

\subsection{Extraction of the Original Sentences}
After ordering near duplicate documents which can potentially include some cases of paraphrasing, we extracted sentences that detected as text matching cases in the \textit{lead} documents by Hamtajoo (i.e., \textit{original} sentences which are highlighted parts of Document A in Figure \ref{fig:5}). \par
The text matching cases are separated from normal texts by a specific HTML tag. So, the pieces of text that detected as cases of plagiarism by the system are extracted from those tags. Since the extracted pieces of texts would include several sentences, we tokenized the whole text into sentences, using Parsivar Persian text pre-processing tool~\cite{DBLP:conf/lrec/MohtajRZA18}. \par

\subsection{Searching for Paraphrased Sentences}
After the extraction of the \textit{original} sentences from the \textit{lead} documents, in this step we search for the \textit{paraphrased} sentences in the \textit{subsequent} documents. For this purpose, we chose the approximate location of the \textit{original} sentence (in the \textit{lead} document) in the \textit{subsequent} documents. In other words, we focused on the position in which the \textit{original} sentence was extracted, in the re-submitted document. Since most of the users try to conceal plagiarism by just paraphrasing those sentences that are detected by the system, we expected to find the \textit{paraphrased} sentences in the similar position. \par
However, since the paraphrased sentence would shift a few characters in the \textit{subsequent} document compared to the \textit{lead} document, we also took into account \textpm 100 characters in the \textit{subsequent} document. It means we looked for the \textit{paraphrased} sentences in span of 100 characters before to 100 characters after the position of the \textit{original} sentences. \par
After choosing a span of \textpm 100 characters in the \textit{subsequent} document, we split the text within the span into sentences. The resulting sentences are the potential \textit{paraphrased} sentences for the \textit{original} sentence. To detect the true \textit{paraphrased} sentence among the potential ones, the \textit{original} sentence and the potential paraphrased sentences have been embedded into vectors, using ParsBERT pre-trained model~\cite{DBLP:journals/npl/FarahaniGFM21}.  ParsBERT is a monolingual BERT for the Persian language, which shows its state-of-the-art performance compared to other architectures and multi-lingual models. We used a Bert based model since it can preserve the semantic representation of the sentences and as a result, can better identify pairs of sentences that are semantically similar. \par
\begin{figure}
\begin{center}
\includegraphics[width=0.36\textwidth]{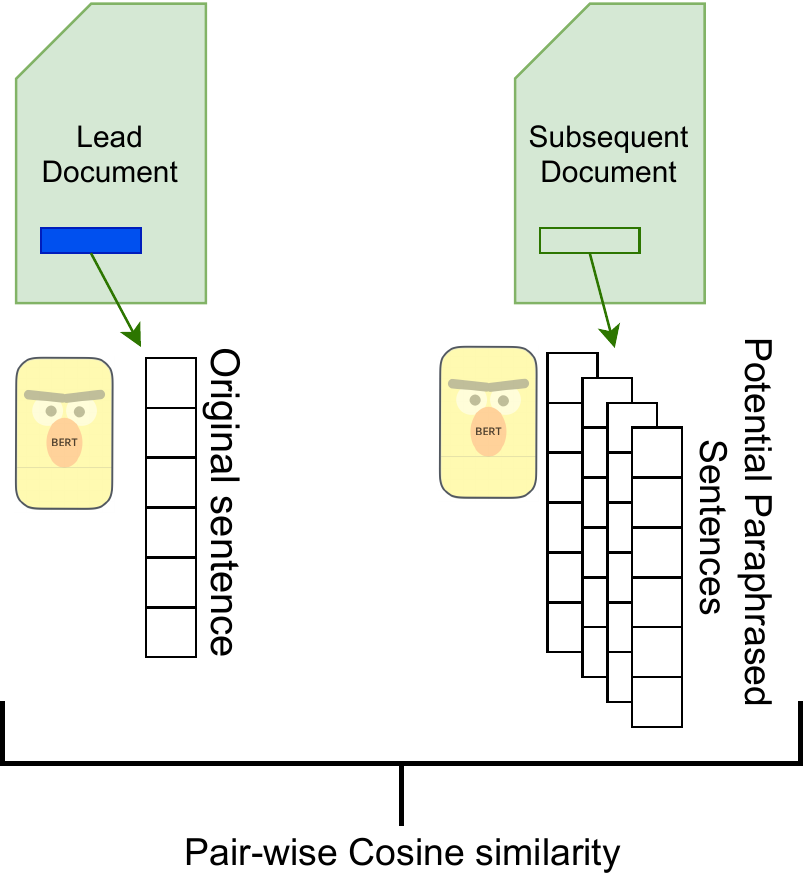}
\caption{Embedding the \textit{original} and potential \textit{paraphrased} sentences into vectors and computing cosine similarity }
\label{fig:6}
\end{center}
\end{figure}

After converting sentences into vector of numbers, the cosine similarity between the \textit{original} sentence and each potential paraphrased sentence is computed as it's shown in Figure \ref{fig:6}. Pairs of sentences with cosine similarity in the range of [0.8 - 1) are extracted as the cases of paraphrasing. This process is repeated for all the \textit{original} sentences in the \textit{lead} document.

\subsection{Post-Processing}
After generating the initial list of pairs of \textit{original} and \textit{paraphrased} sentences in the last step, the low quality candidates are removed in this stage. Since the whole process of extracting pair of paraphrasing sentences have been done automatically, some noises may be added into the list. To remove these noises, we applied a series of heuristics on the list of candidates. For this purpose, we removed those pairs that at least one of the sentences contains at least one of the following conditions:

\begin{itemize}
    \item Sentences that are shorter than 50 characters
    \item Sentences that are not complete (e.g., no subject or verb)
    \item Sentences that are not Persian
\end{itemize}
We used Parsivar~\cite{DBLP:conf/lrec/MohtajRZA18} for Part-of-speech tagging and langdetect\footnote{https://pypi.org/project/langdetect/} to detect the language of the sentences.

\section{Evaluation}
\label{sec:evaluation}
In this section we present the dataset statistics and the validation results.

\subsection{Dataset Statistics}
\begin{figure}[!h]
\begin{center}
\includegraphics[width=0.48\textwidth]{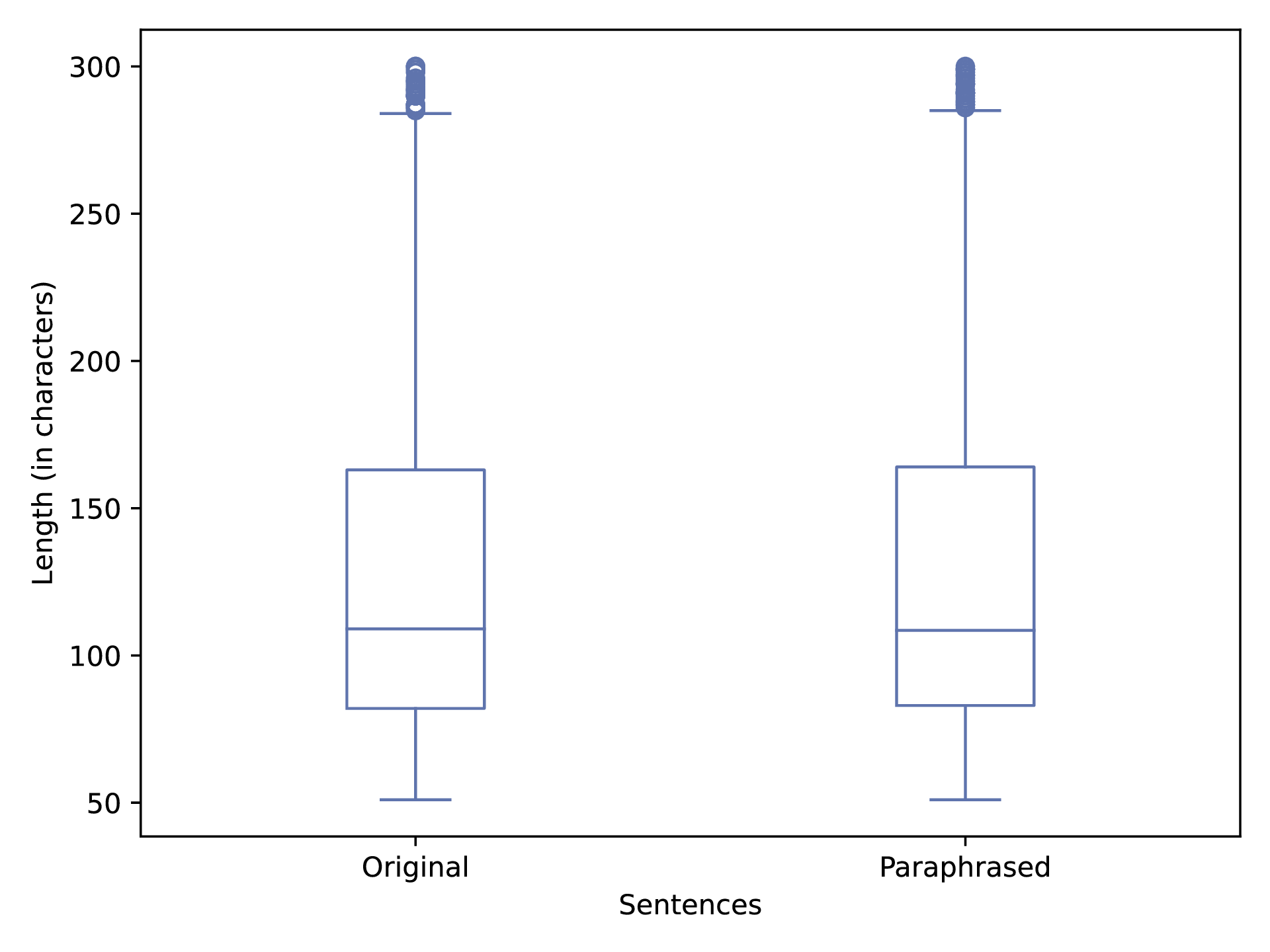}
\caption{Distribution of text length in the original and paraphrased sentences}
\label{fig:7}
\end{center}
\end{figure}

The resulted paraphrased dataset contains \textbf{2446} pair of sentences. The length of sentences vary between almost 50 and 300 characters. The range of lengths in the original and paraphrased sentences are depicted in a box-plot in Figure~\ref{fig:7}. As highlighted in the figure, there is no significant difference between the length of original and paraphrased sentences. 

\begin{figure}[!h]
\begin{center}
\includegraphics[width=0.48\textwidth]{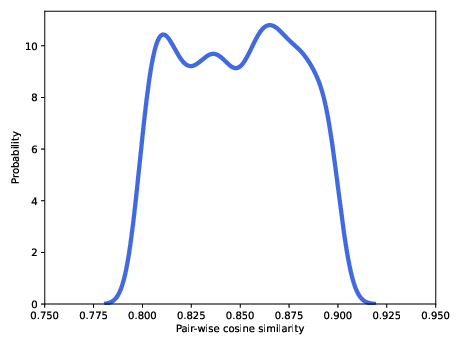}
\caption{Distribution of cosine similarity between the pair of sentences}
\label{fig:8}
\end{center}
\end{figure}

Moreover, the distribution of cosine similarity between the pair of sentences is shown in Figure~\ref{fig:8}. As shown in the figure, similarities vary between 0.8 to 0.92, with a peak on 0.87.

The \textit{PerPaDa} dataset is only available for research purposes by academic researchers. in the corresponding section in Hamtajoo website\footnote{http://hamtajoo.ir/corpus}.

\subsection{Validation Result}
\begin{figure}[!h]
\begin{center}
\includegraphics[width=0.48\textwidth]{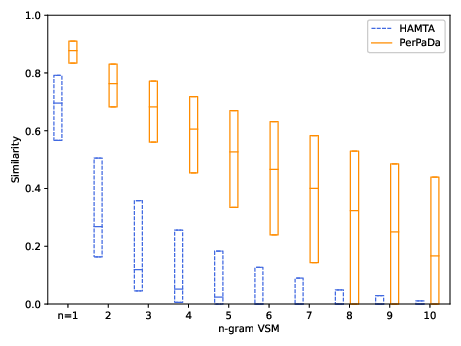}
\caption{Comparison of \textit{PerPaDa} w.r.t. HAMTA: each box plot shows the range of cosine
similarity between the original and the paraphrased sentences.}
\label{fig:9}
\end{center}
\end{figure}

To validate the proposed dataset, we compared the similarity of pairs of sentences in \textit{PerPaDa} with manually paraphrased Persian text in HAMTA, that is a Persian plagiarism detection corpus~\cite{DBLP:journals/jodl/AsghariFMF21},~\cite{DBLP:conf/fire/AsghariMFFRP16a}. HAMTA corpus includes manually and automatically generated paraphrased pieces of text. We took the manually compiled paraphrases from HAMTA to compare with \textit{PerPaDa}. For Comparison, we used the method proposed by Potthast et. al. that includes 10 different retrieval models. Each model is an n-gram vector space model (VSM) where n ranges from 1 to 10 words, employing tf-weighting, and the cosine similarity~\cite{DBLP:conf/coling/PotthastSBR10}. The resulting plot is shown in Figure~\ref{fig:9}.

As it is expected, the overall similarity of the pair of sentences decreases by increasing n from 1 to 10 in both data. It means although the original and paraphrased sentences share a number of terms, these terms usually re-ordered in the paraphrased sentences. However, the pair of sentences in \textit{PerPaDa} tends to have more similarity, comparing to HAMTA. It shows users tend to apply least changes on the original sentence in the implicit crowdsourcing experiment, while they have to make more changes based on pre-defined rules in an explicit crowdsourcing setting.

\section{Conclusion}
\label{sec:conclusion}

In this paper we introduced \textit{PerPaDa}, a Persian Paraphrase Dataset based on an implicit crowdsourcing experiment. The raw data has been collected from Hamtajoo that is a Persian plagiarism detection system. We tried to extract those documents from Hamtajoo in which a user tried to conceal text matching cases by paraphrasing part of text that are detected by system as cases of plagiarism. \par
Based on our validation experiments, the proposed data shows similar results with manually paraphrased corpora. However, an implicitly generated dataset like \textit{PerPaDa} is much more cheaper comparing to the explicitly compiled corpora. It can also better shows the paraphrasing behavior of an ordinary user.

\section{Acknowledgment}
We would like to thank all of the members of ITBM and AIS research groups of ICT research institute for their contribution in developing the Hamtajoo platform.
\section{Bibliographical References}\label{reference}

\bibliographystyle{lrec2022-bib}
\bibliography{perpada}


\end{document}